\documentclass[letterpaper, 10 pt, conference]{ieeeconf}  

\IEEEoverridecommandlockouts                              

\makeatletter
\let\NAT@parse\undefined
\makeatother
\usepackage[numbers]{natbib}

\usepackage{amsmath} 
\usepackage{amssymb}  
\usepackage{bm}
\usepackage{xcolor}
\usepackage{graphicx}
\usepackage{mathtools}
\usepackage{varioref}
\usepackage{hyperref}
\usepackage{siunitx}
\usepackage{tikz}
\usetikzlibrary{shapes}
\usepackage{multirow}
\usepackage{makecell}
\usepackage{algorithm}
\usepackage{algpseudocode}

\usepackage{subcaption}
\usepackage{tablefootnote}
\usepackage{cleveref}

\DeclareMathOperator*{\argmin}{arg\,min}

\definecolor{blue}{HTML}{5d81b5}
\definecolor{orange}{HTML}{e09b24}
\definecolor{green}{HTML}{8eb031}
\definecolor{red}{HTML}{eb6235}
\definecolor{purple}{HTML}{8678b2}
\definecolor{brown}{HTML}{c46e1a}
\definecolor{cyan}{HTML}{5c9dc7}

\hypersetup{
	colorlinks=true,       
	linkcolor=red,        
	citecolor=red,        
	filecolor=red,     
	urlcolor=cyan         
}

\newif\ifcomments
\commentstrue 

\newcommand{\diversity}{\mathrm{Diversity}}

\usepackage{dashrule}
\newcommand{\legendskills}{
	{\color{blue}\hdashrule[.5ex]{0.3em}{.2em}{}}
	{\color{green}\hdashrule[.5ex]{0.3em}{.2em}{}}
	{\color{brown}\hdashrule[.5ex]{0.3em}{.2em}{}}
	{\color{orange}\hdashrule[.5ex]{0.3em}{.2em}{}}
	{\color{red}\hdashrule[.5ex]{0.3em}{.2em}{}}
        {\color{purple}\hdashrule[.5ex]{0.3em}{.2em}{}}
        {\color{brown}\hdashrule[.5ex]{0.3em}{.2em}{}}\quad \footnotesize{Skills} \\
}

\definecolor{alpha_0_h}{HTML}{d52321}
\definecolor{alpha_0_m}{HTML}{fc8161}
\definecolor{alpha_0_l}{HTML}{fedbcb}
\definecolor{alpha_1_h}{HTML}{2b944b}
\definecolor{alpha_1_m}{HTML}{8ed08c}
\definecolor{alpha_1_l}{HTML}{e1f4db}
\definecolor{alpha_2_h}{HTML}{2b7bba}
\definecolor{alpha_2_m}{HTML}{88bedc}
\definecolor{alpha_2_l}{HTML}{dae8f6}

\newcommand{\method}{DOMiNiC}

\newcommand{\norm}[1]{\left\lVert#1\right\rVert}
\newcommand{\innerproduct}[2]{\langle #1, #2 \rangle}

\title{\LARGE \bf
Learning Diverse Skills for Local Navigation under \\ Multi-constraint Optimality
}
\author{Jin Cheng$^{1,2}$, Marin Vlastelica$^{1}$, Pavel Kolev$^{1}$, Chenhao Li$^{1,2}$, Georg Martius$^{1,3}$
\thanks{$^{1}$Max Planck Institute for Intelligent Systems, Germany
        {\tt\small \{firstname.lastname\}@tuebingen.mpg.de}}%
\thanks{$^{2}$ETH Zürich, Switzerland
        {\tt\small \{jicheng, chenhli\}@ethz.ch}}%
\thanks{$^{3}$University of Tübingen, Germany}%
}

\begin{document}
\maketitle
\thispagestyle{empty}
\pagestyle{empty}

\begin{abstract}
Despite many successful applications of data-driven control in robotics, extracting meaningful diverse behaviors remains a challenge.
Typically, task performance needs to be compromised in order to achieve diversity.
In many scenarios, task requirements are specified as a multitude of reward terms, each requiring a different trade-off.  
In this work, we take a constrained optimization viewpoint on the quality-diversity trade-off and show that we can obtain diverse policies while imposing constraints on their value functions which are defined through distinct rewards. 
In line with previous work, further control of the diversity level can be achieved through an attract-repel reward term motivated by the Van der Waals force.
We demonstrate the effectiveness of our method on a local navigation task where a quadruped robot needs to reach the target within a finite horizon.
Finally, our trained policies transfer well to the real 12-DoF quadruped robot, Solo12, and exhibit diverse agile behaviors with successful obstacle traversal.
\end{abstract}



\section{Introduction}
\label{sec:introduction}
Reinforcement Learning (RL) has proven itself as a valuable tool for equipping robotic platforms with a variety of capabilities. 
In the realm of legged systems, RL has achieved commendable success in quadrupedal locomotion control within challenging environments~\citep{lee2020learning, kumar2021rma, miki2022learning}.
However, the ability of RL to provide a range of diverse solutions for the same task still remains a challenging frontier.
The core of our research lies in addressing this limitation. 

Given a set of reward functions describing a particular task, our goal is to train a variety of different skills that solve the same task proficiently.
This essentially formulates a constraint optimization problem as proposed by \citet{zahavy2022discovering}, where the objective is to maximize diversity while satisfying constraints that guarantee that each skill achieves a certain level of cumulative reward in comparison to an expert trained with only task rewards. 

In terms of diversity objectives, a promising approach is based on the online maximization of skill discriminability where skills are represented by latent variables on which a policy is conditioned~\citep{eysenbach2018diversity, hansen2019fast, strouse2021learning}.
While mutual-information-based objectives~\cite{gregor2016variational, sharma2019dynamics} encourage each skill to deviate from the overall average, it does not ensure pairwise skill diversity.
As an alternative, \citet{zahavy2022discovering} proposed a physics-inspired force as a diversity objective measured in a feature space that combines repulsive and attractive terms to allow for different degrees of pairwise skill diversity.
Diverse skills can be obtained using the DOMiNO~\cite{zahavy2022discovering} algorithm that ensures each skill has near-expert performance with respect to the extrinsic reward.

\begin{figure} [t]
    \vspace{0.2cm}
    \centering
    \includegraphics[width=0.75\linewidth]{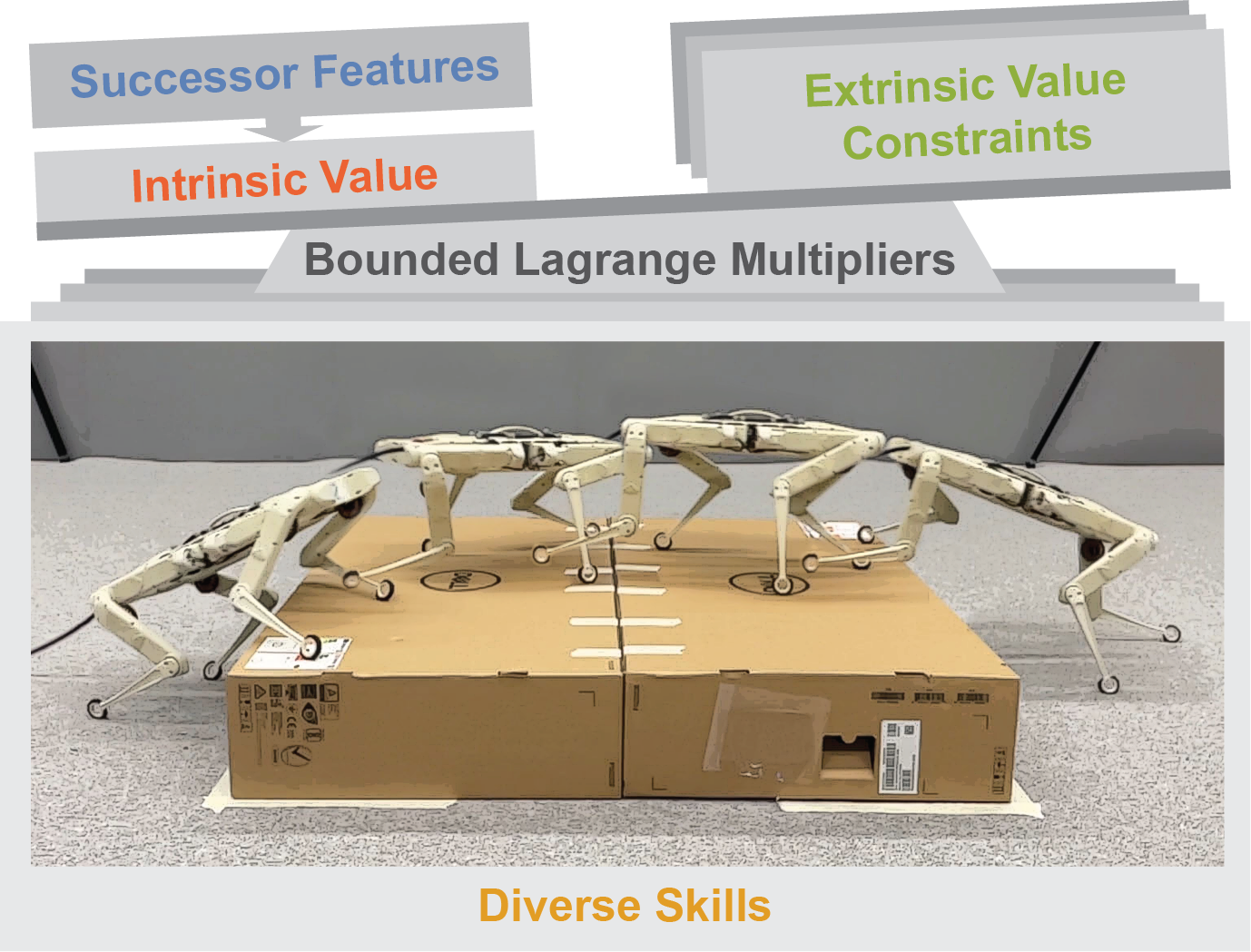}
    \caption{Our framework (\method{}) uses a gradient-based Lagrange method to maximize diversity within a specified set of constraints.
    The learned skills exhibit diverse behaviors in real-world robotic systems. {Project website with videos: \url{https://sites.google.com/view/icra2024-dominic}}}
    \label{fig:teaser}
    \vspace{-0.5cm}
\end{figure}

In this study, we introduce Diversity Optimization under Multiple Near-optimal Constraints (\method{}), an adaptive extension to the DOMiNO~\cite{zahavy2022discovering} framework. \method{} is tailored for addressing intricate challenges within the realm of real-world legged robotics when dealing with tasks featured by multiple objectives in a local navigation task within a terrain of obstacles. 
\method{} is particularly effective in environments characterized by a wide variety of constraints, including important facets such as task-based rewards, safety-oriented regularization, and discretionary auxiliary rewards -- all with different requirements on how much they can be sacrificed.

Our contribution is threefold: 
\textbf{i)} We propose \method{}, an RL algorithm that achieves quality-diversity balance with an arbitrary number of constraints;
\textbf{ii)} We conduct an extensive study of the controllability aspects and the trade-offs between maximizing skill diversity and satisfying constraints on cumulative rewards in the domain of quadruped locomotion;
\textbf{iii)} Our experimental evaluation validates the effectiveness of \method{} in transferring a skill-conditioned policy trained in simulation to Solo12, a 12-DoF quadruped robot operating in a real-world environment.

\section{Related Work}
\label{sec:related_work}
In this section, we focus on two aspects that are closely related to our work: RL-based control of quadruped robots and unsupervised skill discovery. 

RL-based methods have recently shown their prominent capability in controlling legged systems~\citep{tan2018sim, hwangbo2019learning, rudin2022learning}. 
Trained in simulation, RL policies can achieve great robustness in tracking velocity commands over challenging terrains~\citep{lee2020learning, kumar2021rma, miki2022learning}. 
However, velocity-commanded policies often converge to a single behavior exhibiting trotting gaits in the case of quadruped systems. 
Efforts have been made to use RL to achieve diverse behaviors by hierarchical control, imitation, and unsupervised skill discovery. 
By formulating locomotion as a position-based local navigation task, ~\citet{rudin2022advanced} has recently shown agile behaviors emerging from RL such as climbing boxes and crossing gaps. These behaviors have been recently used as motion priors in a hierarchical control structure for long-horizon navigation tasks~\citet{hoeller2023anymal} that preserves the diversity. 
Despite the impressive results, different priors require behavior-specific reward design, which needs a substantial amount of reward engineering to balance the behavior and regularization for all priors.
Alternatively, combining skill-conditioned policy with imitation objective can largely reduce the extensive reward shaping. 
Imitating reference trajectories has shown the ability to generate agile behavior for quadruped robots such as dog-like hopping~\cite{peng2020learning}, backflipping~\cite{li2023learning} and walking with two feet~\cite{fuchioka2023opt}. 
\citet{kang2023rl+} proposed to condition the locomotion policy on four phase variables in combination with imitating trajectories from model-based controllers to achieve different gait patterns on quadruped robots. 
Despite the skillful locomotion results, imitation-based methods often require prior knowledge of the robotic system and the resulting behavior is affected by the quality of the reference trajectories.

As an alternative, unsupervised skill discovery has recently gained research attention in the RL community, which is often related to maximizing the skill difference across policies that are conditioned on latent variables. 
The intrinsic objective can be incorporated into online training to discover diverse behaviors, and most recently~\citet{vlastelica2023diverse} also proposed a Fenchel-duality approach for offline skill discovery.
Mutual-information-based methods quantify the shared information between latent variables and the historical states by a skill-conditioned policy and maximize the mutual information between the skill and states to obtain distinct behaviors from each other~\cite{eysenbach2018diversity, sharma2019dynamics}, which has been shown to extract diverse behaviors successfully in combination with other rewards~\cite{kumar2020one, li2023versatile}. 
Alternative to the mutual-information objective, diversity can also be measured by the Euclidean distance in the state or feature space~\cite{park2021lipschitz, park2023controllability}. 
\citet{li2023versatile} combined the imitation objective with unsupervised skill discovery to extract diverse skills from unlabeled offline demonstrations. 

Despite the interesting motions discovered by these methods, acquiring meaningful and task-related behavior still remains challenging due to the need to balance quality and diversity carefully.
Recently, \citet{zahavy2022discovering} proposed DOMiNO to combine unsupervised skill discovery with Constrained Markov Decision Processes (CMDPs)~\citep{altman1999constrained, cmdpblog} to ensure near-expert task performance as well as diversity in behaviors. 
CMDPs, which are a crucial part of RL with implications for safety and risk aversion, are first used to achieve quality-diversity balance.

As an extension to DOMiNO from~\citet{zahavy2022discovering}, \method{} focuses on combining CMDPs to unsupervised skill discovery for real robotic systems. Specifically, we focused on a similar scenario as~\citet{rudin2022advanced} in training locomotion policies for quadruped robots with diverse behaviors to accomplish the local navigation task.

\section{Preliminaries}
\label{sec:preliminaries}
The core idea of DOMiNO~\cite{zahavy2022discovering} is to utilize the CMDP formulation to maximize two reward components: extrinsic and intrinsic rewards, which are balanced through Lagrange multipliers to achieve the quality-diversity trade-off. 

In the general RL setup, an agent interacts with an environment to maximize the cumulative discounted reward. 
From the definition of Markov decision processes (MDPs)~\citep{puterman2014markov}, an initial state $s_0$ is sampled from a state distribution $\rho(s_0)$, then at each time step $t$, the agent applies an action $a_t$ according to a policy $\pi(a_t|s_t)$ given a state $s_t$, and receives from the environment a reward $r_t\sim R(s_t, a_t)$ and a next state $s_{t+1}\sim P(s_{t+1} | s_t, a_t)$.
The performance metric can be written as $v_{\pi} = (1 - \gamma) \mathbb{E} [\sum_{t=0}^{\infty} \gamma ^t r_t]$ and the state-action occupancy is defined as $d_\pi(s, a) = (1 - \gamma) \mathbb{E} [\sum_{t=0}^{\infty} \gamma ^t P_{\pi}(s_t=s)\pi(a|s)]$. The RL objective can be rewritten as maximizing a function of the occupancy measure $\max_{d_\pi \in \mathcal{K}}\innerproduct{d_{\pi}}{r}$, where $\innerproduct{d_{\pi}}{r}=\sum_{s, a}d_\pi(s, a)r(s, a)$ denotes the inner product and $\mathcal{K}$ is the set of admissible distributions~\citep{ZahavyODS21}. 

\subsection{Constrained Markov Decision Process}

\citet{zahavy2022discovering} studied the CMDP formulation, which seeks to compute a set of policies $\Pi^n = \{\pi^z\}_{z = 1}^n$ that satisfy
\begin{equation}
\max_{\Pi^n} \ \diversity(\Pi^n) \ \text{s.t.} \ \innerproduct{d_{\pi}}{r_e} \geq \alpha v^*_e,\quad \forall \pi \in \Pi^n,
\label{eq:constrained_mdp}
\end{equation}
where $r_e$ and $v_e^*$ correspond to the extrinsic reward and optimal extrinsic value.
Intuitively, it computes a set of diverse policies while maintaining a certain level of extrinsic optimality specified by the optimality ratio $\alpha \in[0, 1]$.

As shown in previous work~\citep{ZahavyODS21}, convex diversity objectives can be optimized by solving a sequence of standard RL problems, each with an intrinsic reward equal to the gradient of the objective evaluated at a state-action occupancy $d_\pi$ of the current iteration:
\begin{equation}
    r_i^z = \nabla_{d^z_\pi}\diversity(d^1_\pi, ..., d^n_\pi), \ \quad\forall z.
\label{eq:intrinsic_reward}
\end{equation}

The CMDP in~\cref{eq:constrained_mdp} can be formulated into an RL problem by utilizing the Lagrange multiplier $\lambda\geq0$ to balance the extrinsic and intrinsic reward~\cite{borkar2005actor}
\begin{equation}
    r^z = r_e + \lambda^z r_i^z, \ \quad\forall z,
\label{eq:reward_combination}
\end{equation}
where the Lagrange multiplier is optimized for the loss
\begin{equation}
    \mathcal{L}_{\lambda} = \sum_{z = 1}^n \lambda^z(\alpha v^*_e - v_e^z).
\label{eq:lagrange_loss}
\end{equation}
Intuitively, the Lagrange multiplier increases when the constraint is fulfilled and decreases otherwise.
The practical implementation considers an extrinsic and intrinsic advantage $a_e, a_i^z$ coupled with bounded Lagrange multipliers~\citep{StookeAA20}, applying Sigmoid activation to unbounded variable $\mu\in\mathbb{R}^{|Z|}$
\begin{equation}
    a^z = \sigma(\mu^z) a_e + (1 - \sigma(\mu^z)) a_i^z, \ \quad\forall z.
\label{eq:advantage_combination}
\end{equation}

\subsection{Diversity Measures}\label{subsec:DiversityMeasures}

Based on a distance measure from \citet{abbeel2004apprenticeship}, \citet{zahavy2022discovering} modeled the diversity objective as the maximization of the minimum squared $\ell_2$ distance between feature expectations of different skills, namely

\begin{equation}\label{eq:rep_force}
    \max_{d^1_\pi, ..., d^n_\pi} \ 0.5 \sum_{z=1}^n \min_{k \neq z}\norm{\psi^z - \psi^k}^2_2.
\end{equation}
More specifically, given a feature mapping $\phi : \mathcal{S}\rightarrow \mathbb{R}^n$, the feature expectations are defined by $\psi^z=\mathbb{E}_{d_{\pi}^{z}(s)}[\phi(s)]$.
Furthermore, \citet{zahavy2022discovering} introduced a physically inspired objective based on Van der Waals (VDW) force, and considered the following optimization objective 
\begin{equation}\label{eq:vdw_force}
    \max_{d^1_\pi, ..., d^n_\pi} \ 0.5 \sum_{z=1}^n \ell_z^2 - 0.2 (\ell_z^5/\ell_0^3), 
\end{equation}
where $\ell_z = \min_{k \neq z}\norm{\psi^z - \psi^k}_2$, which allows the level of diversity to be controlled by $\ell_0$.
When the features are in close proximity $\ell_i<\ell_0$, the repulsive force dominates, whereas when $\ell_i>\ell_0$ the attractive force prevails. 

\begin{figure*}
    \vspace{0.2cm}
    \centering
    \includegraphics[width=0.7\linewidth]{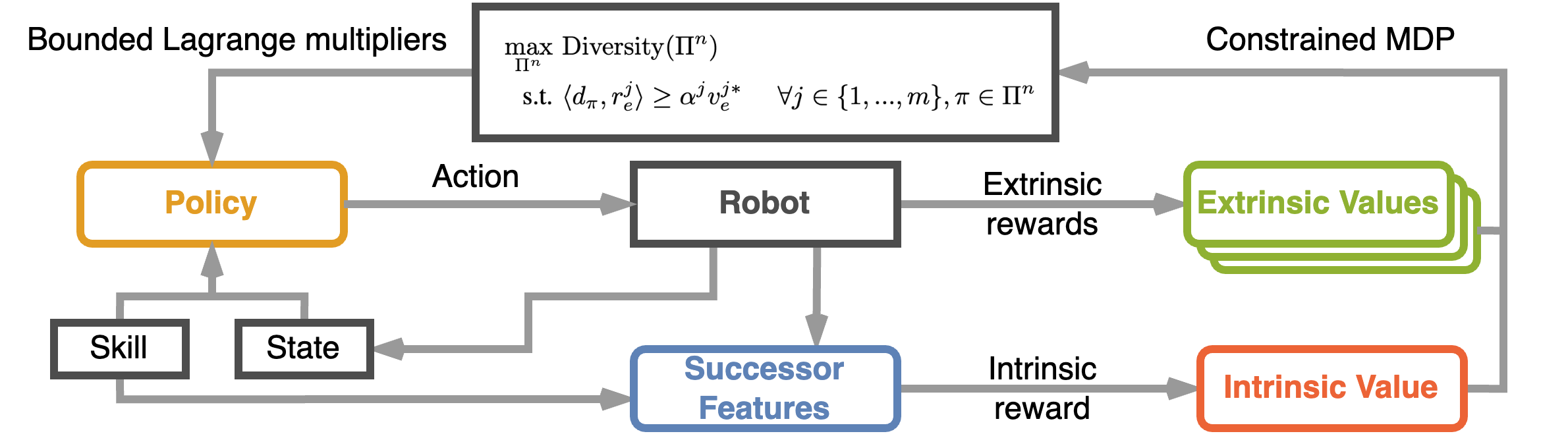}
    \caption{The \method{} training scheme. 
    We collect samples in simulation and fit the extrinsic values for updating the Lagrange multipliers and an intrinsic value function based on \cref{eq:rep_rew} (repulsive) or \cref{eq:vdw_rew} (VDW) used for measuring diversity. 
    These are combined into an aggregate advantage term in \cref{eq:multi_advantage_combination}, ensuring that intrinsic reward is maximized only after all constraints are satisfied, i.e., to train a skill-conditioned policy that solves the CMDP.}
    \label{fig:method}
    \vspace{-0.5cm}
\end{figure*}

\section{Method}
\label{sec:method}

The DOMiNO~\cite{zahavy2022discovering} framework utilizes a single scalar extrinsic value as a metric to assess the proficiency of learned skills.
However, this methodology faces challenges when dealing with tasks characterized by multiple objectives, as it lacks clarity in discerning which particular objective may undergo compromise.
In real-world robotics scenarios, this problem is particularly pronounced. 
Learned skills are expected to adhere to specific human-designed regularizer rewards that include safety or risk-averse criteria, which are critical for enhancing the successful transfer of learned skills from simulation to the physical environment.
Moreover, the spectrum of different optimality considerations cannot be adequately formulated within the single constraint MDP formulation in \cref{eq:constrained_mdp}. 

To address these problems, we present Diversity Optimization under Multiple Near-optimal Constraints (\method{}) to extend the capacity of this framework to multi-constraint optimization scenarios.
More specifically, we categorize the rewards into $m$ different groups.
Each constraint group $j$ has an associated reward $r_e^j$ and an optimality ratio $\alpha^j$.
We consider the following formulation:
\begin{equation}
\begin{split}
    \max_{\Pi^n} \ &\diversity(\Pi^n) \\
    \text{s.t.} \ & \innerproduct{d_{\pi}}{r_e^j} \geq \alpha^j v_e^{j*} \ \quad\forall j\in \{1, ..., m\}, \pi \in \Pi^n,
\end{split}
\label{eq:multi_constrained_mdp}
\end{equation}
where $r_e^j$ is the extrinsic reward and $v_e^{j*}$ is the optimal value of group $j$. 
This multi-constraint formulation allows fine-grained control over different constraint groups, via the parameters $\{\alpha^j\}_{j=1}^{m}$.
An overview of our framework is shown in~\cref{fig:method}.

We incorporate the advantage-based method with bounded Lagrange multipliers as discussed in~\Cref{sec:preliminaries}.
To ensure that the intrinsic reward is maximized only after all constraint groups have been satisfied, we introduce the following aggregate advantage term
\begin{equation}\label{eq:multi_advantage_combination}
    a^{z}=\Big(1-\max_{j}{\sigma(\mu^{j,z})}\Big)a_{i}^{z}+\sum_{j}\sigma(\mu^{j,z})a_{e}^{j},\quad\forall z,
\end{equation}
where $a_e^j$ is the extrinsic advantage for reward group $j$, $a_i^z$ is the intrinsic advantage of skill $z$, and $\sigma(\mu^{j,z})$ is the bounded Lagrange multiplier for constraint $j$ and skill $z$.
Note that the aggregate advantage $a^z\rightarrow a_i^z$ when $\sigma(\mu^{j,z})\rightarrow 0$ for all constraint groups, i.e., when all constraints are satisfied.

The Lagrange multipliers are updated according to the following loss function, which is designed to guarantee the satisfaction of all constraint groups:
\begin{equation}
    \mathcal{L}_\mu = \sum_{z = 1}^n \sum_{j = 1}^m \ \mathop{\mathbb{E}}_{v_e^{j,z}}\left[\mu^{j,z}(\alpha^j v_e^{j*} - v_e^{j,z})\right].
    \label{eq:lm_loss}
\end{equation}

To compute the feature expectations $\psi^z$ in \Cref{subsec:DiversityMeasures}, we use the state-conditioned Successor Features (SFs) proposed by \citet{barreto2017successor}, which decouple the environment dynamics from rewards and facilitate knowledge transfer across tasks.
The SFs of a policy $\pi$ evaluated at a state $s$ are given by
$\psi^{z}(s)=\mathbb{E}_{z}\big[\sum_{i=0}^{\infty}\gamma^{t}\phi(s_{t})\ |\ s_{0}=s\big]$.

Our algorithm relies on the following two properties:
i) feature expectations satisfy
\begin{equation}\label{eq:FE_SFs}
\psi^{z}=\mathop{\mathbb{E}}_{\rho(s_0)}[\psi^{z}(s_0)],
\end{equation}
where $\rho(s_0)$ is the initial state distribution; and
ii) SFs $\psi^{z}(s)$ can be trained by a learning process similar to training a value function, using Temporal Difference (TD) updates~\citep{Sutton1998}, which minimizes the loss
\begin{equation}\label{eq:sf_loss}
    \mathcal{L}_\psi = \sum_{z=1}^n \mathop{\mathbb{E}}_{\pi^z}\norm{\phi(s) + \gamma \psi^z(s') - \psi^z(s)}_2^2.
\end{equation}
At each time step, the intrinsic reward $r_i$ is computed from the learned SFs $\psi(s)$ either by the repulsive force in~\cref{eq:rep_force}
\begin{equation}\label{eq:rep_rew}
    r_{i}^{z}(s)=\langle\phi(s),\psi^{z}-\psi^{z^{\star}}\rangle
\end{equation}
or by the VDW force in~\cref{eq:vdw_force}
\begin{equation}\label{eq:vdw_rew}
    r_{i}^{z}(s)=(1-(\ell_{z}/\ell_{0})^{3})\langle\phi(s),\psi^{z}-\psi^{z^{\star}}\rangle,
\end{equation}
where $z^{\star} = \argmin_{k\neq z}\norm{\psi^{z}-\psi^{k}}_{2}$. 

To estimate the optimal extrinsic values $v_e^{j*}$ for all $j$, we pretrain an expert policy by setting $\sigma(\mu^{j}) = 1$ such that the corresponding policy $\pi^*$ only optimize extrinsic reward. 
During training diverse policies, we maintain moving averages of extrinsic values and use them in the Lagrangian loss $\mathcal{L}_\mu$, as discussed by~\citet{zahavy2022discovering}:

\begin{equation}
    \Bar{v}_e^{j,z} = \alpha^{\text{avg}}\Bar{v}_e^{j,z} + (1 - \alpha^{\text{avg}})v_e^{j,z}, \ \quad\forall j, z.
    \label{eq:moving_avg}
\end{equation}

Instead of concatenating the one-hot encoder of discrete skills to the input as in previous work, we use randomly initialized layer masks for all skill-conditioned neural networks, including the policy, value functions, and SFs. 
Before training, a binary mask of the same size as the hidden dimensions is sampled and fixed for each skill.
During training, the mask activation is used to set the output of the corresponding neural units to zero. 
Using these masks ensures that individual skills retain distinct features, mitigating interference and promoting skill diversity within the neural network architecture.
The mask sampling probability is chosen to balance the independence and overlap of neural units between skills.

In contrast to prior work, our approach employs an on-policy training paradigm, drawing inspiration from Proximal Policy Optimization (PPO)~\cite{schulman2017proximal}.
This choice leads us to utilize Generalized Advantage Estimation (GAE)~\cite{schulman2015high} as our preferred method, as opposed to the V-trace technique~\cite{espeholt2018impala} typically associated with off-policy frameworks.
We also introduce a ``warm-start'' phase to warm up all skills to near-expert level, by pre-training them solely with extrinsic advantages.
The complete pseudocode is given in~\cref{alg:x}.

\begin{algorithm}
\caption{\method{}}
\label{alg:x}
\small
\begin{algorithmic}[1]
\Require $\pi$: Policy network, $v_i, \{v^j_e\}_{j=1}^m$: intrinsic and extrinsic value networks, $\psi(s)$: SFs network, $\mu$: Lagrange multipliers, $\psi^{z}$: feature expectations, $\Bar{v}_e$: Moving average of extrinsic values. 
\State Initialize networks, Lagrange multipliers, rollout buffer $\mathcal{B}$
\For{learning iterations = 1,2, ...}
\State sample latent skill variable $z\sim p_z$
\For{time step = 0,1,2, ...}
\State collect transition $(s, a, s', \phi(s), \{r_e^j\}_{j=1}^m)$ with $\pi^z$
\State compute $r_i$ using either~\cref{eq:rep_rew} or~\cref{eq:vdw_rew}
\State fill rollout buffer $\mathcal{B}$ with $(s, a, s', \phi(s), \{r_e^j\}_{j=1}^m, r_i)$
\EndFor
\State compute TD targets for value update,
\State estimate advantages $\{a_e^j\}_{j=1}^m, a^z_i$ for policy update
\For{policy learning epoch = 1,2, ...}
\State sample transition mini-batches $b \sim \mathcal{B}$ 
\State compute aggregate advantage using $\sigma(\mu)$ in~\cref{eq:multi_advantage_combination}
\State update $\pi$ and $v_i, \{v^j_e\}_{j=1}^m$ with PPO objective
\State update SFs network $\psi(s)$ by the loss~\cref{eq:sf_loss}
\State update feature expectations $\psi^{z}$ by~\cref{eq:FE_SFs}
\State update moving averages $\Bar{v}_s$ using~\cref{eq:moving_avg}
\If{not warm start}
\State update $\mu$ by the loss~\cref{eq:lm_loss}
\EndIf
\EndFor
\EndFor
\end{algorithmic}
\end{algorithm}

\subsection{Quadruped Locomotion and Local Navigation}
\label{subsec:pos-based_quad_loco}
We demonstrate our method on the task of quadruped locomotion and local navigation in a position-based framework proposed by~\citet{rudin2022advanced}, where the robot needs to navigate through an environment of randomly positioned boxes of different dimensions to reach a specified target position and orientation within a finite time horizon.

The observations of the policy include base linear and angular velocity, joint position and velocity, gravity vector projected in the base frame, and the height measurement sampled around the robot. 
Random noise is added to these observations to simulate hardware sensor noise. 
In addition, the policy observes the previous actions, the three-dimensional target position in the base frame, the target yaw angle difference from the current base yaw angle, and a time indicator. 
The action of the policy is the target joint positions which will be taken by a PD controller and transformed into joint torques. 

We group extrinsic reward terms into three categories: 
task $r_t$, regularizer $r_r$, and style $r_s$. 
The task reward is computed from the distance to the target at the end of an episode, so the robot is free to choose the trajectory and gait as long as it reaches the target. 

The task reward $r_t$ consists of the rewards to track the target position and target orientation.
\begin{equation}\label{eq:task-reward}
    r_t = r_{\mathrm{pos}} + r_{\mathrm{yaw}}.
\end{equation}
The position tracking reward is defined using the target position in the base frame $x^*_b$
\begin{equation}\label{eq:position-tracking-reward}
    r_{\mathrm{pos}} = (1+\norm{x^*_b })^{-1}.
\end{equation}
The orientation tracking reward is defined using the target yaw angle from the current yaw angle $\theta^*_b$
\begin{equation}\label{eq:orientation-tracking-reward}
    r_{\mathrm{yaw}} = (1+\norm{\theta^*_b})^{-1} \cdot (\norm{x^*_b} \leq 0.25),
\end{equation}
which is non-zero when the target position is tracked well.

The regularizer reward $r_r$ includes different components to regularize the behavior as well as to ease sim-to-real transfer
\begin{equation}\label{eq:regularizer-reward}
    r_r = r_{\Dot{a}} \cdot r_{c} \cdot r_{\tau} \cdot r_{g} \cdot r_{st},
\end{equation}
where $r_{\Dot{a}}, r_{c}, r_{\tau}$ are used to regularize action rate, non-feet contact and large torques, $r_{g}$ is to regularize large roll and pitch angles of the base by comparing the projected gravity vector with the global one, and $r_{st}$ is defined to encourage the robots to have a minimum velocity (not stall) when the target position is far. 
Each of them is mapped by an exponential function $r_x = \exp\{-\norm{x/\sigma_x}^2\}$, where $x$ is the corresponding value and $\sigma_x$ is a scaling factor. 

The style reward $r_s$ comprises rewards aimed at guiding robots to adopt a specific style based on prior knowledge. 
Notably, these rewards are not essential to task completion
\begin{equation}\label{eq:sacrifiable-reward}
    r_s = r_{ft} \cdot r_{mt} \cdot r_{q},
\end{equation}
where $r_{ft}$ assigns a higher reward when robots face the target, $r_{mt}$ motivates robots to move towards the target, and $r_{q}$ keeps all joint angles close to the default ones. 

By setting large optimality ratios $\alpha_t, \alpha_r$ for the task and regularizer groups, we intuitively seek skills that can both track the target and have regularized motion that can be transferred to the real system.
At the same time, we can diversify the locomotion style by setting different optimality ratios $\alpha_s$ to the style group.  
In principle, the fixed feature mapping $\phi(s)$ can be chosen arbitrarily, but a careful selection using either human or learned expertise can lead to favorable outcomes.
We highlight the importance of this choice by presenting the result of two different mappings in~\Cref{sec:experiments}.

\subsection{Training Details}

\begin{figure}[H]
\centering
\begin{subfigure}[b]{0.9\linewidth}
   \includegraphics[width=1\linewidth]{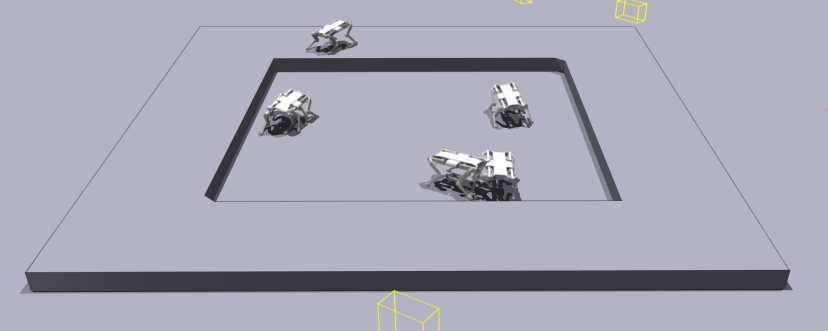}
\end{subfigure}
\vspace{0.1cm}
\\
\centering
\begin{subfigure}[b]{0.9\linewidth}
   \includegraphics[width=1\linewidth]{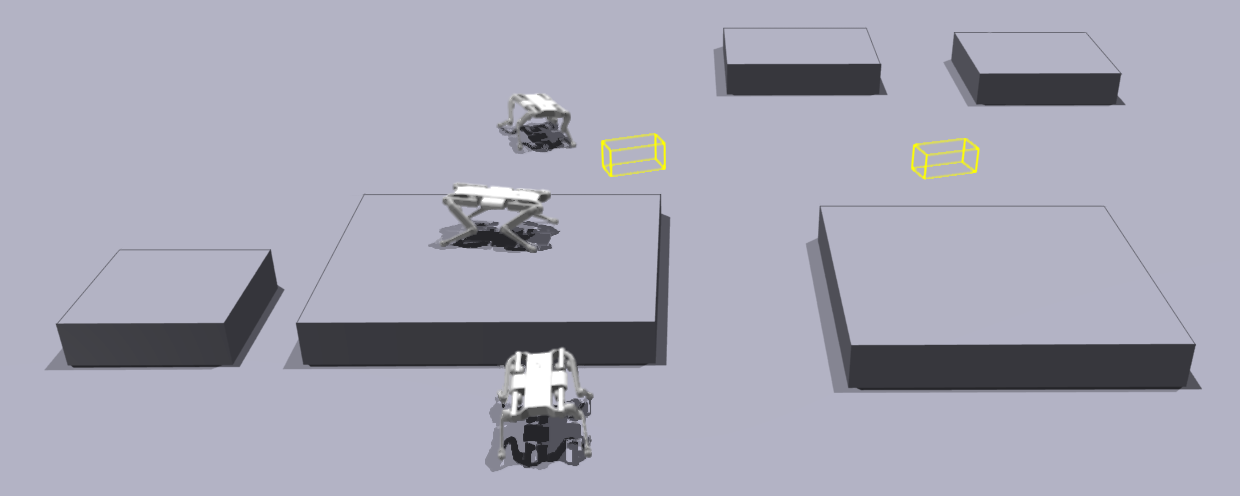}
\end{subfigure}
\label{fig:diversity}
\caption{Terrain types used for training, \textit{top}: barrier, \textit{bottom}: box. Target positions are depicted as yellow boxes. }
\vspace{-0.2cm}
\label{fig:terrain}
\end{figure}

All robots are simultaneously trained on terrains with uniformly sampled boxes, characterized by random sizes within the intervals [0.8, 2.0] meters for length, and [0.0, 0.2] meters for height.
Prior to this, a curriculum of barrier terrains is implemented to warm-start skills with the ability to track targets and to climb on and off boxes of varying heights, which is consistent with the warm-start phase discussed earlier. 
The two terrain types are shown in~\cref{fig:terrain}.
The curriculum involving increased heights in barrier terrains is similar to the approach of~\citet{rudin2022learning}.

In addition to the warm-start phase, we also randomize the mass and friction coefficients, simulate random pushes on robots, and apply a 15 ms actuator delay to bridge the gap between simulation and reality.

\section{Experiments}
\label{sec:experiments}

We evaluate our method on Solo12, an open-source quadruped platform~\cite{leziart2021implementation} with 12 degrees of freedom both in simulation and on real hardware. 
We train a set of policies for 2000 iterations consisting of 48 simulation steps with 4096 parallel environments in Isaac Gym~\cite{makoviychuk2021isaac} including 800 iterations of warm-starting, which takes about 3 hours using GeForce RTX 3080Ti GPU. 
Similar to the previous work of~\citet{rudin2022advanced}, we fix the episode length to 6 seconds and give the task reward $r_t$ in the last second. 

\subsection{Skill Discovery in Local Navigation}
\label{subsec:skill_disc_in_local_navi}
In our first experiment, we show in simulation that diverse skills can be learned to successfully navigate to the target position in the presence of obstacles, while achieving a good balance between task, regularization, and style with optimality ratios $[\alpha^t, \alpha^r, \alpha^s] = [0.9, 0.8, 0.7]$.
We choose $\phi(s)$ to map a state to a base velocity direction expressed in the base frame, with the aim of achieving diverse orientations towards the target.
In addition, diversity can be controlled using the intrinsic objective derived from the VDW force in~\cref{eq:vdw_rew}. 
We empirically validate the learned skills in front of a square obstacle with a size of 1.4 and a height of 0.2 meters.

\begin{figure}[H]
    \centering
    \legendskills{}
    \vspace{0.2cm}
    \begin{subfigure}[t]{0.48\linewidth}
        \centering
        \includegraphics[width=1.0\linewidth]{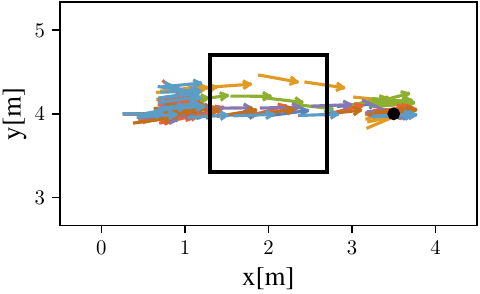}
    \end{subfigure}%
    ~ 
    \begin{subfigure}[t]{0.48\linewidth}
        \centering
        \includegraphics[width=1.0\linewidth]{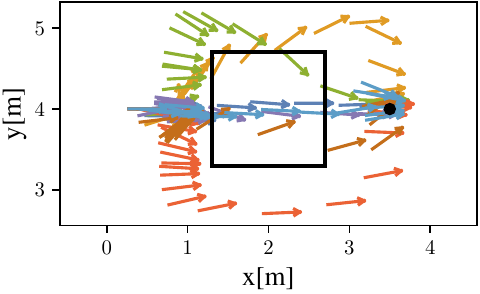}
    \end{subfigure}
\caption{The top-down view of diverse trajectories of successful obstacle traversal obtained with different $\ell_0$ values (\textit{left}: small $\ell_0$, \textit{right}: large $\ell_0$) under the combination $[\alpha^t, \alpha^r, \alpha^s] = [0.9, 0.8, 0.7]$. 
Arrows indicate the yaw angle of the robot at trajectory points.
The black squares are the visualization of the box obstacle and the black circles are the target positions that the robots have to reach.}
\label{fig:diversity_sim}
\vspace{-0.2cm}
\end{figure}

In the top-down view of~\cref{fig:diversity_sim}, the learned skills exhibit different base velocity directions while moving towards the target and different strategies, including detours and jumping on the box, to overcome the obstacle.
For small values of $\ell_0$, forcing low diversity, all learned skills converge to the ``shortest path'' solution, which is characterized by reaching the target position by jumping on the box and then jumping down to the target position on the ground.

\begin{figure*}[t!]
    \vspace{0.2cm}
    \centering
    \includegraphics[width=0.9\linewidth]{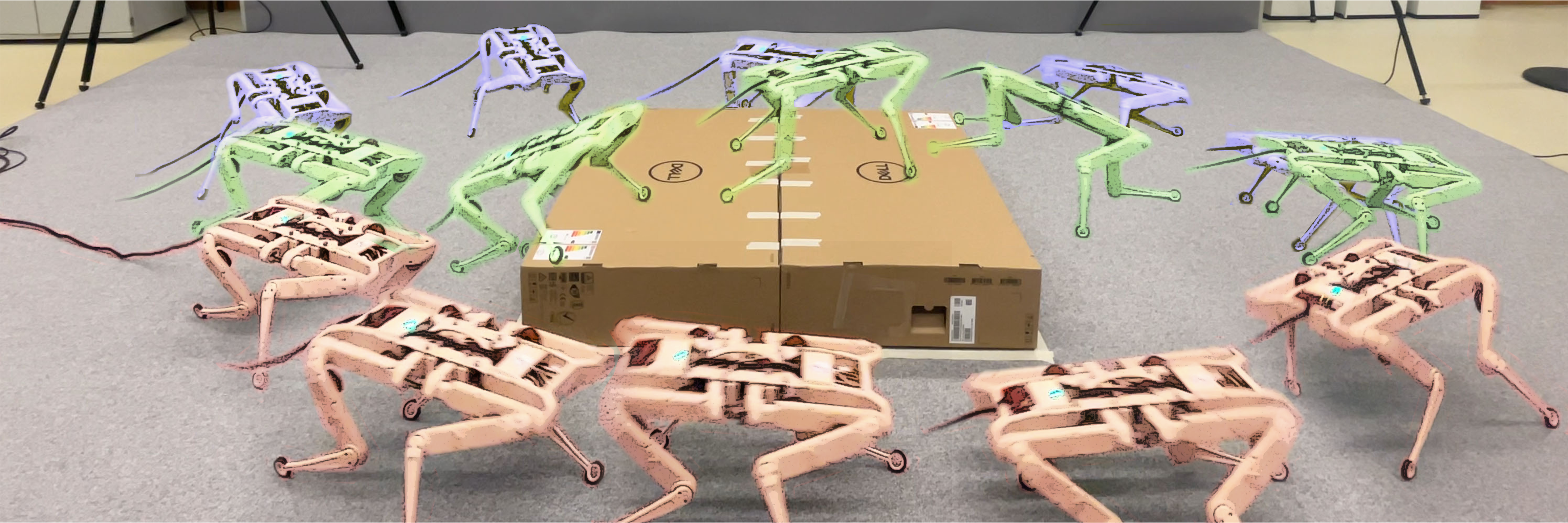}
    \caption{Obstacle experiment on hardware, we observe that the extracted skills explore different options in solving the obstacle. We have skills that go over the obstacle, to the right or to the left in different styles. The green skill is the one closest to the expert, which never takes detours around the box.}
    \label{fig:hardware}
    \vspace{-0.5cm}
\end{figure*}

\subsection{Quality-Diversity Balance}
\label{subsec:q_d_balance}

\begin{figure}[t]
    \vspace{0.2cm}
    \centering
    \begin{subfigure}[b]{0.48\linewidth}
        \centering
        \includegraphics[width=\textwidth]{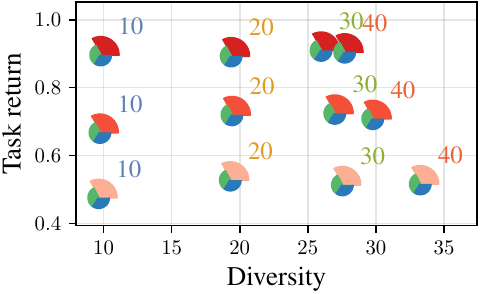}  
    \end{subfigure}
    \hfill
    \begin{subfigure}[b]{0.48\linewidth}  
        \centering 
        \includegraphics[width=\textwidth]{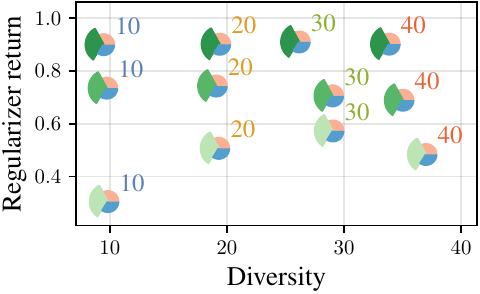}
    \end{subfigure}
    \vskip\baselineskip
    \begin{subfigure}[b]{0.48\linewidth}   
        \centering 
        \includegraphics[width=\textwidth]{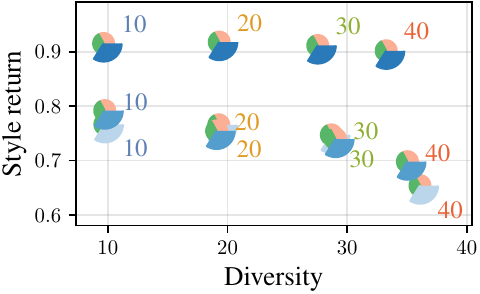}
    \end{subfigure}
    \hfill
    \begin{subfigure}[b]{0.48\linewidth}   
        \centering 
        \vspace{-0.5cm}
        \includegraphics[width=\textwidth]{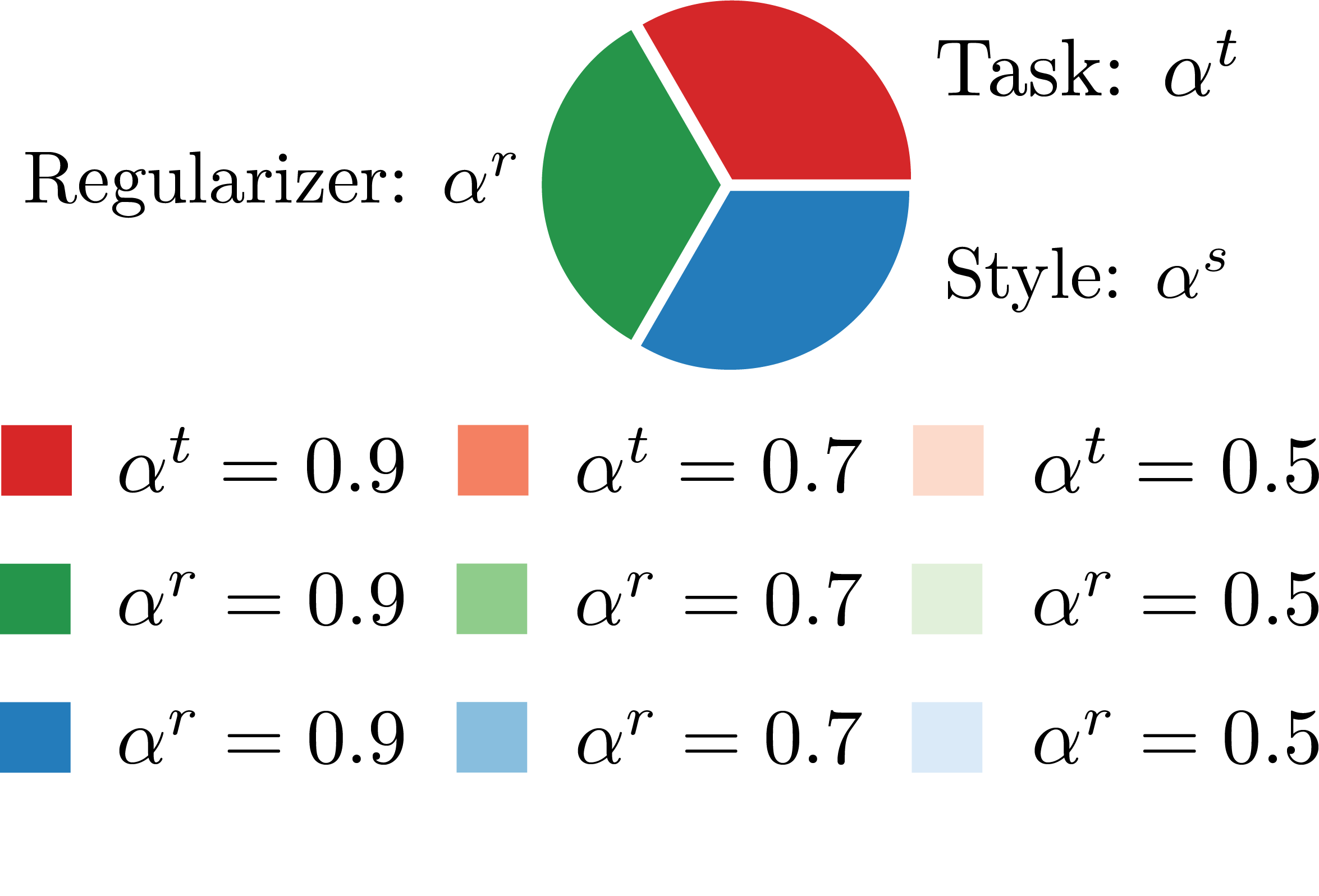}
    \end{subfigure}
\caption{The controllability over quality and diversity through different values of optimality ratios $[\alpha^t, \alpha^r, \alpha^s]$ and $\ell_0$ from the VDW force objective. 
The results of a grid search over three different values $\{0.5, 0.7, 0.9\}$ for $[\alpha^t, \alpha^r, \alpha^s]$ and four values $\{10, 20, 30, 40\}$ for $\ell_0$ are shown as scattered pie plots. 
The colors on the three sectors represent different values for each optimality ratio, and the $\ell_0$ levels are shown as annotations at the points. 
For each figure, we fix two of the three optimality ratios and plot the return that corresponds to the varying term on the vertical axis: \textit{top left}: $[*, 0.7, 0.9]$, \textit{top right}: $[0.5, *, 0.7]$, \textit{bottom left}: $[0.5, 0.7, *]$.}
\vspace{-0.5cm}
\label{fig:vdw}
\end{figure}

In the second experiment, we perform an extensive grid search on different combinations of optimality ratios $[\alpha^t, \alpha^r, \alpha^s]$ and different values of $\ell_0$ in the VDW intrinsic reward in~\cref{eq:vdw_rew} to evaluate their influence on diversity. 
With the $\phi(s)$ mapping the state to the base velocity and joint angles, we seek skills with diverse joint configurations while accomplish the task. 
The results are shown in~\cref{fig:vdw}. 
For the demonstration videos in simulation, we refer the interested readers to the project website. 

The intrinsic reward in~\cref{eq:vdw_rew}, allows us to set a desired level of diversity by $\ell_0$.
On the horizontal axis, the diversity is plotted by measuring the mean of the closest distance between the feature expectations $\psi^z$ evaluated on a uniform initial state distribution $\rho(s_0)$.
Optimality ratios $[\alpha^t, \alpha^r, \alpha^s]$ give us the budget of how much extrinsic reward we can sacrifice for a gain in diversity. 
The vertical axis shows the percentage of returns achieved relative to the expert.

Overall, we find good controllability of the diversity via $\ell_0$ as well as of the quality of the behavior via the optimality ratios $[\alpha^t, \alpha^r, \alpha^s]$.
It is important to emphasize that with looser constraints (smaller $\alpha$) we gain more diversity as shown in all three plots in~\cref{fig:vdw}. 

In addition to demonstrating controllability, several insights can be derived from~\cref{fig:vdw}.
First of all, we notice that the controllability of different reward groups are not entirely independent of each other. 
Imposing task and regularizer constraints leads to an over-satisfaction of the style constraint as shown from the bottom left plot in~\cref{fig:vdw}. 
Second, it is more difficult to control the optimality of the regularizer group. 
We hypothesize that different reward terms in the multiplicative structure of~\cref{eq:regularizer-reward} might influence each other, making the regularizer value harder to maintain at a moderate level.

In summary, we have achieved good controllability over both quality and diversity. 
However, managing the intricate interactions among reward groups and their sub-components remains a promising direction for future research.

\subsection{Hardware Deployment}
We deploy our trained policy from~\cref{subsec:skill_disc_in_local_navi} on the real robot as shown in~\cref{fig:hardware}, where the robot manages to choose diverse trajectories to reach the target behind a box obstacle with a width of 1.4 meters and height of 0.18 meters. 
Different skills extract diverse ways to traverse the obstacle by either jumping onto it or taking a detour around it from the left or the right.
The policy deployed on hardware was trained with large optimality ratios for the task and regularizer to ensure good task performance and to fulfill the action smoothness required on a real system. In addition, intrinsic reward with repulsive force in~\cref{eq:rep_rew} was used to maximize the achieved diversity. 
For estimating the robot base state, a Vicon motion capture system is used to provide the base position and orientation at 100 Hz, and the velocity is calculated based on the finite difference.
The position of the box obstacle is fixed in the global frame, so the height scan is created based on the absolute position of the robot. 


\section{Conclusion}
\label{sec:conclusion}
We propose \method{}, a framework that effectively controls the trade-off between diversity and extrinsic rewards with multiple constraints by leveraging the CMDP formulation and incorporating the Van der Waals force as an intrinsic objective. 
We successfully train policies with diverse skills for Solo12, a 12-DoF quadruped robot tasked with locomotion and local navigation. 
The learned behaviors exhibit various successful obstacle traversal strategies in the real-world robotic system. 
Furthermore, the satisfaction of each constraint group contributes to the achievement of natural and diverse behaviors, emphasizing the significance of our proposed multi-constraint diversity optimization framework.




\bibliographystyle{IEEEtranN}
\bibliography{main.bib}

\end{document}